\def\year{2016}
\documentclass[letterpaper]{article}
\usepackage{aaai}
\usepackage{times}
\usepackage{helvet}
\usepackage{courier}

\usepackage{multirow}
\usepackage{epstopdf}
\usepackage{pgfplots}
\usepackage{algorithmic}
\usepackage{algorithm}
\usepackage{subfigure}
\usepackage{amsmath}
\usepackage{amssymb}  
\usepackage{booktabs}
\usepackage{tabularx}

\usepackage{listings}
\usepackage{xcolor}
\usepackage{textcomp}
\definecolor{listinggray}{gray}{0.9}
\definecolor{lbcolor}{rgb}{0.9,0.9,0.9}
\makeatletter

\floatstyle{ruled}
\newfloat{algorithm}{tbp}{loa}
\providecommand{\algorithmname}{Algorithm}
\floatname{algorithm}{\protect\algorithmname}

\usepackage{algorithm}
\usepackage{algorithmic}

\makeatother

\begin{document}
%
\title{Accelerated Distance Computation with Encoding Tree for High Dimensional Data}
\author{
  Liu Shicong, Shao Junru, Lu Hongtao\\
  \texttt{\{artheru, yz\_sjr, htlu\}@sjtu.edu.cn Shanghai Jiaotong University}
}
\maketitle
\begin{abstract}
\begin{quote}
We propose a novel distance to calculate distance between high dimensional vector pairs, utilizing vector quantization generated encodings. 
Vector quantization based methods are successful in handling large scale high dimensional data. 
These methods compress vectors into short encodings, and allow efficient distance computation 
between an uncompressed vector and compressed dataset without decompressing explicitly. 
However for large datasets, these distance computing methods perform excessive computations. 
We avoid excessive computations by storing the encodings on an Encoding Tree(E-Tree), 
interestingly the memory consumption is also lowered. 
We also propose Encoding Forest(E-Forest) to further lower the computation cost. 
E-Tree and E-Forest is compatible with various existing quantization-based methods. 
We show by experiments our methods speed-up distance computing for high dimensional data drastically, 
and various existing algorithms can benefit from our methods.
\end{quote}
\end{abstract}

\section{Introduction}

The rapid development of the Internet in the recent years brings explosive growth of information online. 
Researchers have been developing methods utilizing such huge amount of data 
for machine learning, information retrieval, computer vision, etc. 
Because the majority of large-scale datasets consists of high-dimensional data, 
there is an increasing requirement for efficient basic operations like evaluating distance and computing scalar product.

Product Quantization (PQ)\cite{pq} is a typical method for fast distance computation/scalar product on high-dimensional data. 
PQ compress high-dimensional data into short encodings, and is able to evaluate distances or 
scalar product between uncompressed and compressed vectors without explicit decompression. 
Given a $d$-dimensional dataset, PQ compress a dataset by first splitting the vector dimensions into $M$ groups,
then quantize each dimension group separately to generate $M$ codebooks containing $K$ codewords (each codeword has $d/M$ dimensions). 
Finally we pick one codeword form each codebook to encode an input vector. 
The compressed vector has $M$ parts, each part occupies $\log_{2}K$ bits. 
An encoded vector is approximated (decompressed) by the concatenation of $M$ codewords assigned.

Computing distances between $N$ pairs of PQ compressed vectors and an uncompressed vector $\mathbf{x}$ can be efficiently done in $O(MN)$ time, 
via a smart use of lookup tables. It is introduced as Asymmetric Distance Computing (ADC) in \cite{pq}. 
One can easily extend the idea to allow efficient scalar product computation\cite{du2014inner}, etc. 
PQ enables efficient Approximate Nearest Neighbor search, where PQ achieves favorable memory / speed vs 
accuracy trade-offs against several competitive methods including Hashing based schemes and Tree based schemes\cite{opq}, \cite{ck}. 
Researchers also developed various quantization methods motivated by Product Quantization. 
e.g. Tree Quantization\cite{babenko2015tree}, Composite Quantization\cite{composite}, 
Cartesian K-means\cite{ck}, Additive Quantization\cite{babenko2014additive}, etc, 
to further lower the quantization error. 

\textbf{Existing problem:} Though ADC is efficient compared to directly computing the distances, 
it still does excessive computations. Existing vector quantization methods simply store the encodings sequentially in the memory, 
and exhaustively perform ADC to compute the approximate distance. However in any quantized dataset, 
many encodings share the same prefixes. These prefixes are repeatedly computed with ADC, they also take up excessive memory.

\textbf{Our contribution:} In this paper, we propose Encoding Tree(E-Tree) to lower the memory consumption and 
speedup the distance computation for encodings generated with vector quantization methods. 
An E-Tree is a compact version of prefix tree with the nodes having only one leaf child recursively merged. 

We propose Hierarchical Memory Structure for Encoding Tree which is designed for efficient depth first traversal 
and allow accelerated distance computation. To perform accelerated distance computation, 
we maintain a very short "partial" ADC results, and depth-first traverse the tree. 
The accelerated distance computation is cache friendly and easily paralleled as it sequentially access the memory. 
Interestingly, with Hierarchical Memory Structure, we're able to speed up distance computation as well as lower the memory consumption. 
For further speed up one can generate an Encoding Forest by generating multiple E-Trees on different parts of the encodings, 
at a slight cost of memory consumption.

As a method for fast distance computation, E-Tree/E-Forest are totally compatible with various existing quantization methods 
by simply substitute ADC with E-Tree/E-Forest for distance computation. 
E-Forest achieves up to \textbf{111.7\%} speedup compared to the naive ADC, 
and E-Tree lower the memory consumption by \textbf{12.5\%}. 
E-Tree/E-Forest can accelerate various related algorithms significantly, 
e.g. Locally Optimized Product Quantization by \textbf{74\%}, and IVFADC by \textbf{81\%}. 
Applications relying on efficient distance computation could greatly benefit from our methods.

\section{Related Work}
Vector Quantization is commonly applied on high-dimensional data for efficiently manipulating the data 
like computing distances between vectors. It essentially maps a vector to a codeword, 
and use the codeword to approximate the original vector. Take Product Quantization as an example, 
it first decompose the original data space as the Cartesian Product of $M$ disjoint lower dimensional subspaces, 
and learn $M$ codebooks $\mathbf{C}_m=\{\mathbf{c}_m(1),\cdots, \mathbf{c}_m(K)\}, m=1,\cdots,M$ for each subspace. 
Then we encode a vector $\mathbf{x}$ with $\mathbf{C}_m$ on the corresponding dimensions to produce an  $M$-encoding: 
$\mathbf{x} \rightarrow i_1(\mathbf{x}), i_2(\mathbf{x}), \cdots, i_M(\mathbf{x})$. 
Padding the codewords with zero chunks to obtain full dimensional codewords, 
vector $\mathbf{x}$ can be reconstructed as 
$\mathbf{x}\approx \mathbf{c}_1(i_1(\mathbf{x}))+\mathbf{c}_2(i_2(\mathbf{x}))+\cdots+\mathbf{c}_M(i_M(\mathbf{x}))$.

We can perform Asymmetric Distance Computation(ADC) introduced in \cite{pq} 
to compute the distance between a vector and quantized vectors. 
The Euclidean distance between a vector $\mathbf{q}$ and a database vector $\mathbf{x}$ is approximated by:

\begin{equation}
\begin{split}
\lVert\mathbf{q}-\mathbf{x}\rVert^2&\approx\lVert \mathbf{q}-\sum_{m=1}^M \mathbf{c}_m(i_m(\mathbf{x}))\rVert^2\\ 
&=\sum_{m=1}^M\lVert \mathbf{q}-\mathbf{c}_m(i_m(\mathbf{x}))\rVert^2-(m-1)\lVert \mathbf{q}\rVert^2 \\
&\quad+\sum_{i=1}^M\sum_{j=1,j\neq i}^M \mathbf{c}_i(i_i(\mathbf{x}))^\mathrm{T}\mathbf{c}_j(i_j(\mathbf{x}))
\end{split}
\label{Equ}
\end{equation}

ADC allows fast massive distance computation: The first term is computed only once for all vectors 
before the distance computation and is stored in the a \textbf{precomputed distance table}, 
the second term is a constant for all database vectors which can be omitted, 
and the third term is zero for PQ learned codebooks. 
Thus, the approximate distance between $\mathbf{q}$ and a database vector $\mathbf{x}$ 
can be efficiently computed with $M$ table lookups and $M-1$ addition.
One can also easily extend ADC to perform efficient scalar product\cite{du2014inner},
or compute scalar product on kernel space\cite{davis2014asymmetric}.

Researchers also developed various similar quantization based methods to further lower 
the quantization error and allow preciser distance computation. 
Optimized Product Quantization\cite{opq} proposed to rotate the data space for better subspace partition. 
Additive Quantization doesn't decompose data space into orthogonal subspaces, 
instead it uses all dimensions to generate the codebooks. 
While it makes the third term in equation \ref{Equ} to be non-zero, requiring 
additional information to be stored along with the encoded dataset. 
Other methods include Tree Quantization\cite{babenko2015tree}, 
Composite Quantization\cite{composite}, etc, 
they all allow fast distance computation in an ADC-like fashion.

\begin{figure*}[t]
\includegraphics[width=1\linewidth]{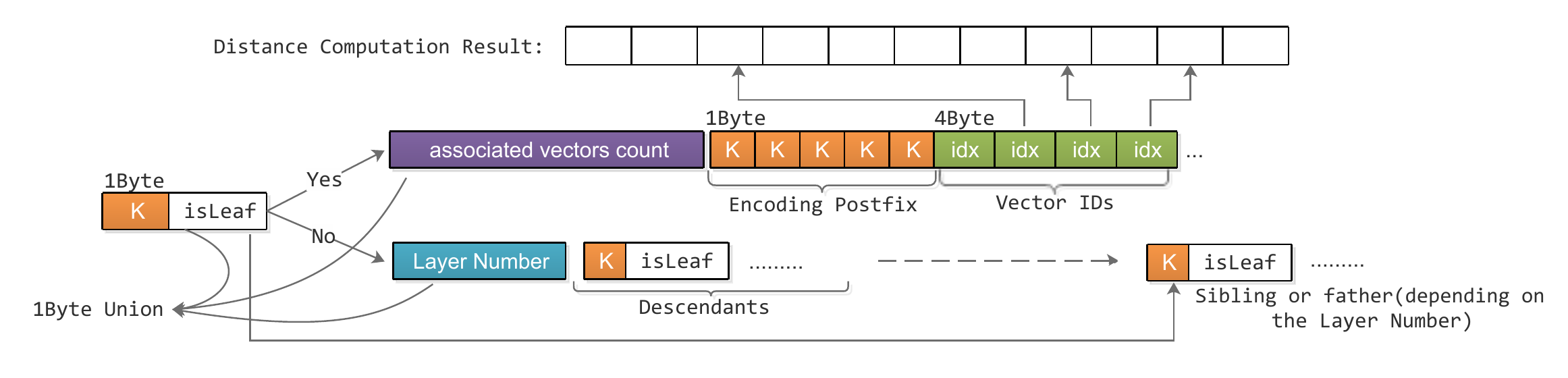}
\caption{Hierarchical Memory Structure layout, the nodes of the encoding tree is stored in depth first traversal sequence, yielding predictable memory access. An internal node takes 2 Bytes; and a leaf node takes $P+2+n'$ Bytes, where $P$ denotes the postfix length and $n'$ denotes the number of associated vectors.}
\label{s}
\end{figure*}
Though ADC is much faster compared to brute force distance computation, 
however, it still makes up the majority consumed time in applications like IVFADC\cite{pq}, 
in large scale SVM training\cite{harchaoui2012large} \cite{lebrun2004svm} 
and other applications involving large scale data. 
For a large scale database contain millions of encoded vectors, 
many encodings have the same prefixes, while these prefixes are repeatedly calculated in ADC. 
Thus a solution is to generate a prefix tree-like structure to discover and avoid excessive computation. 
Interestingly we found such tree also lowers memory consumption. 
We propose Encoding Tree to accelerate distance computation.
\footnote{It can also easily extended to compute inner product or inner product in kernel space. We omit the discussion due to the length limit of the paper.}

\section{Encoding Tree}
\begin{figure}
\includegraphics[width=\linewidth]{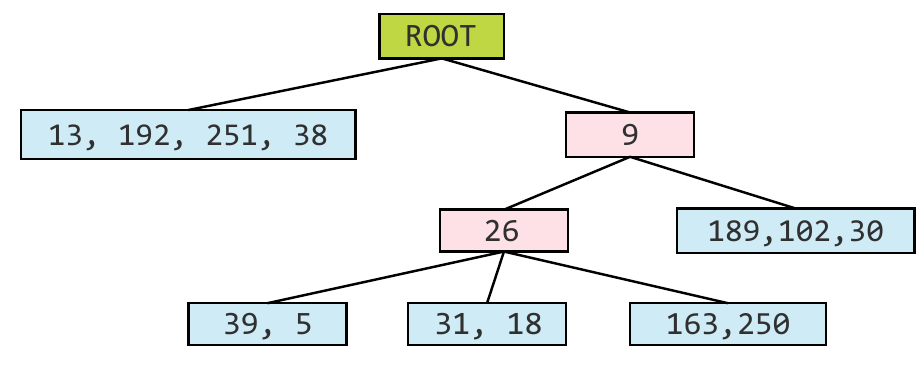}
\caption{An illustrative example of Encoding Tree. Note on this tiny example, we need 25\% less memory access and 13\% less floating point additions to compute the distance compared to ADC implementations. The acceleration is more apparent on large scale datasets.}
\label{et}
\end{figure}
An Encoding Tree is a variant of prefix tree. 
Prefix tree is a standard method for searching and storing strings in scale. 
However prefix tree has not been introduced to handle large scale high-dimensional data occurred to machine learning, 
computer vision, etc, to our knowledge. By generating the encoded vectors, 
one can effectively store encodings of a dataset in a prefix tree, 
in which all the descendants of a node have a common prefix of the encoding associated with that node. 
An illustrative tree structure example is presented in Figure \ref{et}. 
In a prefix tree the common prefix only appears once, 
and the memory it consumes could be saved; 
it also implies that we don't need to calculate the "partial" ADC on this prefix twice. 
Furthermore, in order to achieve higher speed, accuracy and lower memory consumption, 
if one node has only a single leaf descendant, the path from the node to the leaf is compressed into one leaf node.

\subsection{Constructing Encoding Tree}
\begin{algorithm}[H]
\caption{Construction of Encoding Tree}
\label{algo}

\allowdisplaybreaks
\begin{algorithmic}[1]

\REQUIRE $N$ encoded vectors $P_1, \cdots, P_N$, each of which has length of $M$, in lexicographic order
\ENSURE Encoding Tree

\STATE root $\leftarrow$ $P_1$
\STATE lastPath := root
\FOR{each $P_i$}
	\STATE $l$ := Longest Common Prefix(lastPath, $P$)
	\STATE Merge lastPath\{$l+1 \cdots M $\}
	\STATE Create nodes: lastPath\{$l$\} $\leftarrow P\{l+1\} \leftarrow P\{l+2\} \cdots\leftarrow P\{M\}$
	\STATE lastPath\{$l+1 \sim M$\}:=$P \{l+1 \sim M\}$
	\STATE associate $i$-th vector to lastPath\{$M$\}
\ENDFOR

\RETURN root

\end{algorithmic}
\end{algorithm}
To construct the Encoding tree, a straightforward solution is to directly adopt an existing implementation of prefix tree library 
and compress the tree to achieve minimal memory consumption. However, these implementations of general purpose prefix tree are 
still too massive for encoding tree with excessive dynamic arrays and pointers. 
Memory consumption is a critical problem in our algorithm, 
because we have to store all the encodings in memory. 
In addition, dynamic arrays and pointers are not friendly to extensive computation. 
Therefore, a memory efficient and computation friendly approach to maintain the encoding tree is in urgent need. 

If the encodings are in order, we can efficiently generate the Encoding Tree without the use of dynamic array and excessive pointers. 
An in place sort of the encoded dataset can be done efficiently with existing libraries. 
Then we adopt the algorithm presented in \textbf{Algorithm \ref{algo}} to generate the Encoding Tree. 
We first allocate enough memory for the tree, then the tree can simply grow linearly without memory fragments. 
The time complexity of generating the tree alone is $O(MN)$ for an dataset containing $N$ encodings with $M$ chunks. 
Taking in-place sort of the encoded dataset into consideration, 
the total preparation time is $O(MNlogN)$. 
The calculation time for constructing the encoding tree is much smaller than the time costly encoding phase, 
which is $O(dKN)$ for encoding with PQ, or $O(dKN+d^2)$ for OPQ.

\subsection{Hierarchical Memory Structure}
We propose Hierarchical Memory Structure for the above algorithm, 
and present the corresponding accelerated asymmetric distance computation methods in this section. 
An illustration of the Hierarchical Memory Structure is presented in Figure \ref{s}. 
The Hierarchical Memory Structure store nodes in the depth-first traversal sequence 
and is thus actually "flat" to allow parallelism and predictable memory access, 
which is crucial in practice for High Performance Computing(HPC). 
We briefly introduce the role of each field:
\begin{itemize}
\item Field \textbf{K} is for storing a single encoding chunk.
\item Field \textbf{Union} is a branch indicator and stores meta-data of a node. 
The least significant bit indicates if this node is a leaf node or an internal node. 
For a leaf node, the rest of the bits are used for storing associated vectors number; 
and for an internal node, they are used to store depth of a internal node.
\item Field \textbf{idx} indicates the associated vectors' IDs of a node. 
This field only occur when the node is a leaf node.
\end{itemize}
As a depth first traversal sequence, Hierarchical Memory Structure can be separately constructed in different memory chunks, 
and concatenate them to form the whole structure. Thus Algorithm \ref{algo} can be effectively accelerated with various parallelism methods.
The memory consumption of Hierarchical Memory Structure depends on the number of internal nodes $L_1$ 
and the number of leaf nodes $L_2$ existing on the encoding tree. An internal node requires 2 Bytes; 
while a leaf node takes up $2+P$ Bytes, denoting $P$ as the average leaf node postfix length . 
For each dataset vector there is a corresponding ID stored in a leaf node, taking $4N$ Bytes in total. 
The total memory consumption is $4N+2(L_1+L_2)+PL_2$ Bytes.

\begin{figure*}[t]
\includegraphics[width=1\linewidth]{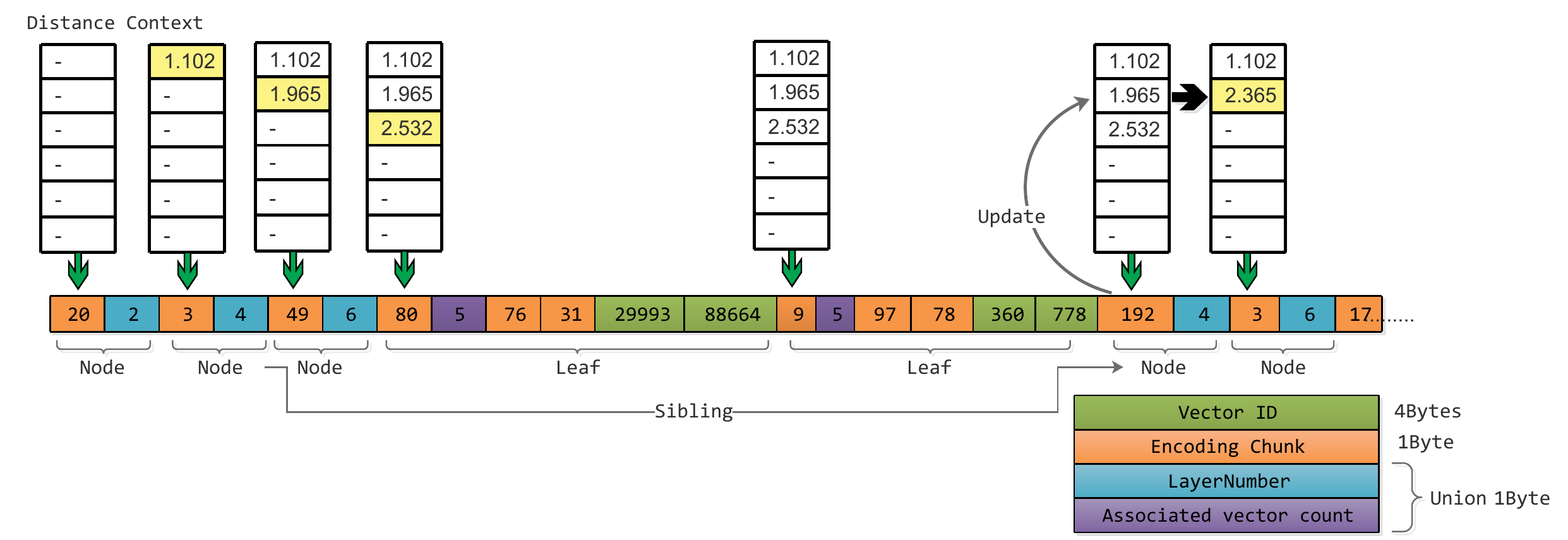}
\caption{To perform distance computation on Hierarchical Memory Structure, one can sequentially access the memory to depth first traverse the tree, and perform "partial" ADC on the current node and store the result on Distance Context Table to avoid excessive computations.}
\label{demo}
\end{figure*}
\subsection{Distance Computation with Hierarchical Memory Structure}
On the distance computation phase, we depth-first traverse the tree(sequentially read the memory in essence), and perform a "partial" ADC on every node. We present a pseudo code in C fashion to elaborate the distance computation:
\begin{lstlisting}
float DistanceContext[M]
Node* pointer=&root
int   currentLayer=0

do{
	distance=DistanceContext[currentLayer]+Precomputed[pointer->K]
	if (pointer->isLeaf){
		int PostfixLength=M-currentLayer
		for (int i=0; i<PostfixLength; ++i)
			distance+=Precomputed[pointer->PostFix[i]]
		OutputResult()
		pointer+=2+PostfixLength+pointer->AssociatedVectorCount
	}
	else{
		currentLayer=pointer->LayerNumber
		DistanceContext[currentLayer]=DistanceContext[currentLayer-1]+Precomputed[pointer->K]
		pointer+=2
	}
}
\end{lstlisting}

We maintain a short \textbf{Distance Context} to store currently performed "partial" ADC result. 
The \textbf{Distance Context} is updated whenever we visit an internal node. 
We output the computation result to a preallocated array for collecting the distance. 
We illustrate the procedure of distance computation in Figure \ref{demo} Note 
the construction of encoding tree \textit{always} merge common prefixes, 
thus every time we update the \textbf{Distance Context}, 
we're avoid excessive computation. 
The final calculation time depends on the number of nodes $N'$ existing on the encoding tree, 
and the average leaf node postfix length $P$\footnote{Or $M-D_l$, where $D_l$ denotes average leaf node depth.}. 
Distance computation with Hierarchical Memory Structure require total $O(N'+P)$ computations.

\begin{figure}
\centering
\includegraphics[width=0.8\linewidth]{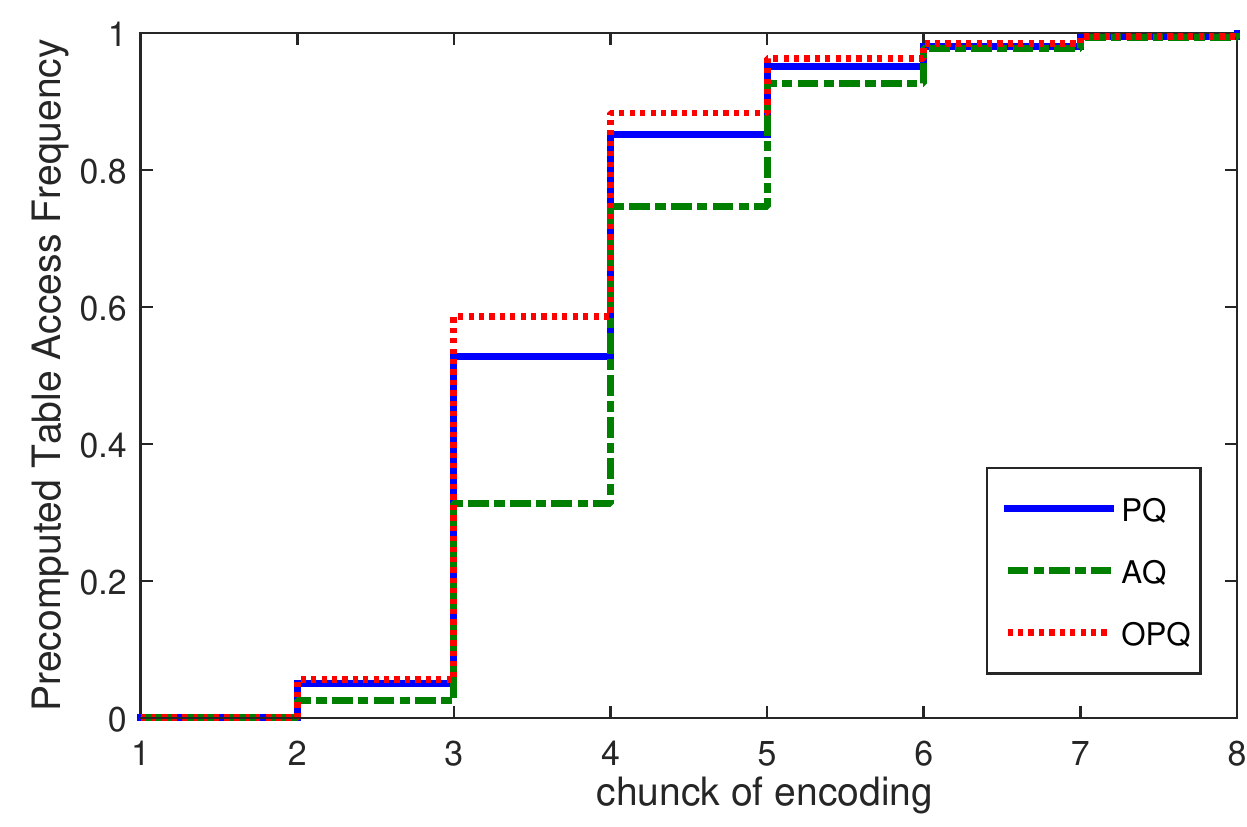}
\caption{The access frequency of different part of Precomputed Distance Table. We used Product Quantization, Optimized Product Quantization and Additive Quantization to produce encoded vectors with $K=256, M=8$, and generate the corresponding Encoding Tree. We perform a traverse of the tree to obtain the access frequency.}
\label{freq}
\end{figure}
Distance computing with Hierarchical Memory Structure requires very few memory. 
The hot spot data is the Distance Context, which can be efficiently stored in the register since the Distance Context only occupy a few Bytes. 
For usual ADC implementation, \textbf{Precomputed distance table} may be too big to fit into a higher level cache, 
while in Hierarchical Memory Structure, only part of the table is frequently accessed so our method is more cache friendly. 
We present the access frequency of each part of the \textbf{Precomputed distance table} in Figure \ref{freq}. 
As a sequential memory accessing method, Hierarchical Memory Model doesn't confuse the prefetch system on CPU/GPU. 
It also allow SIMD instructions available on modern CPU/GPU to further boost the performance.

To sum up, Hierarchical Memory Structure is cache friendly as well as lowers the total computation requirement.

\begin{figure}[!t]
\centering
\subfigure[Number of leaf nodes on the E-Tree/E-forest, by using encoded vectors obtained with different vector quantization methods. ]{
	\includegraphics[width=1\linewidth]{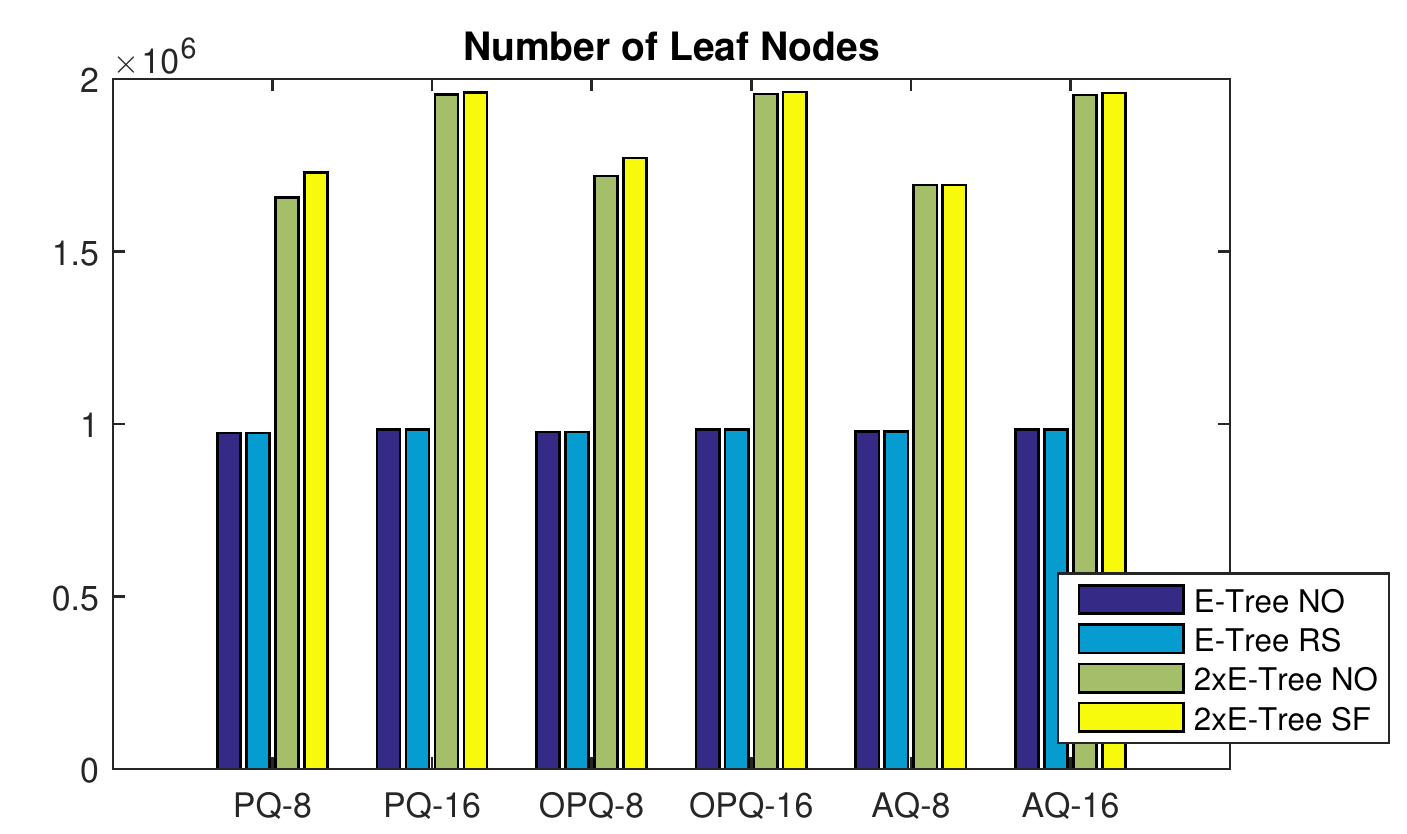}
	\label{leaf}
}
\subfigure[Number of internal nodes on the E-Tree/E-forest.]{
	\includegraphics[width=1\linewidth]{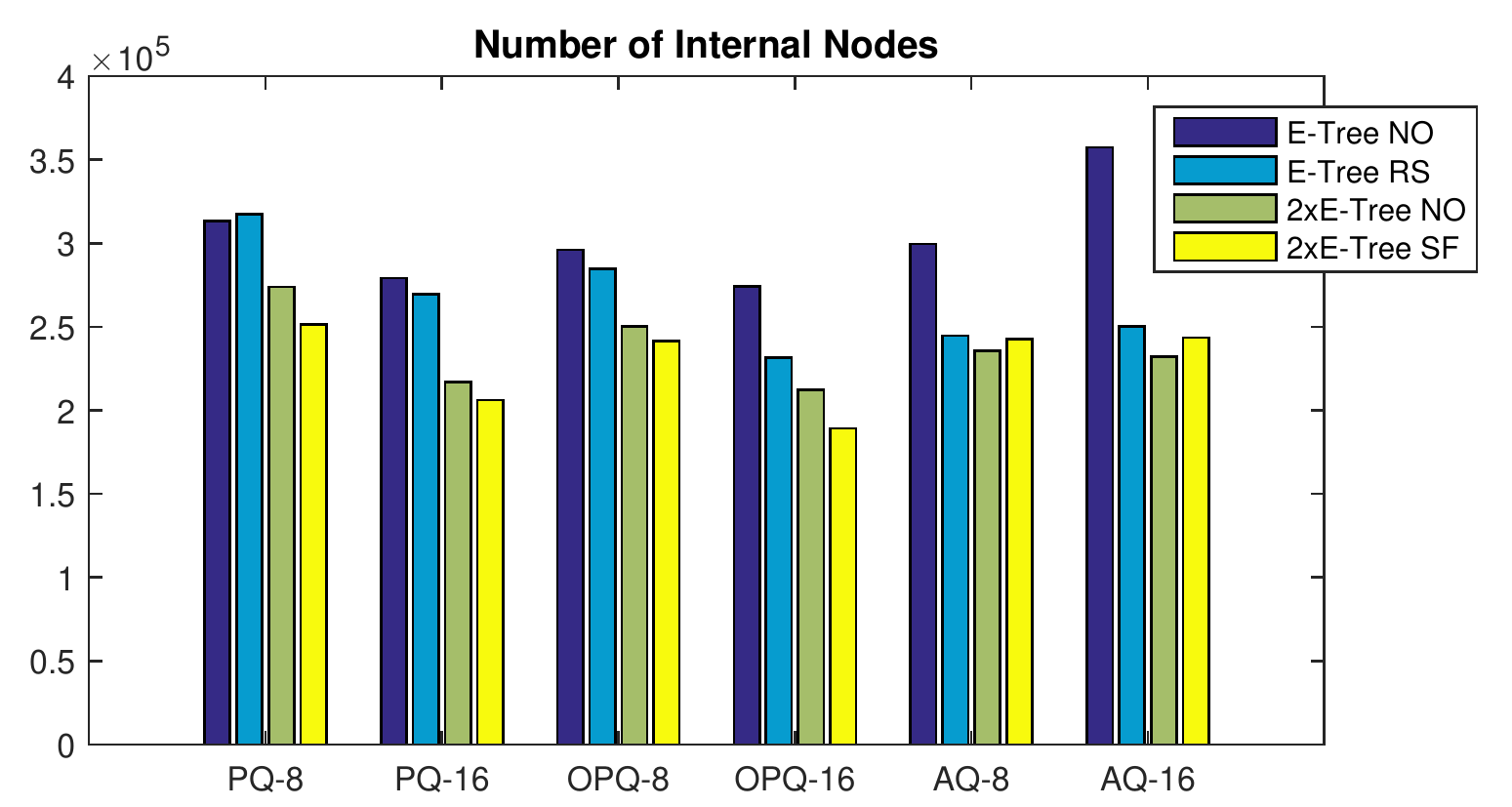}
		\label{internal}
}
\subfigure[Average postfix length of the leaf nodes for E-Tree/E-forest.]{
	\includegraphics[width=1\linewidth]{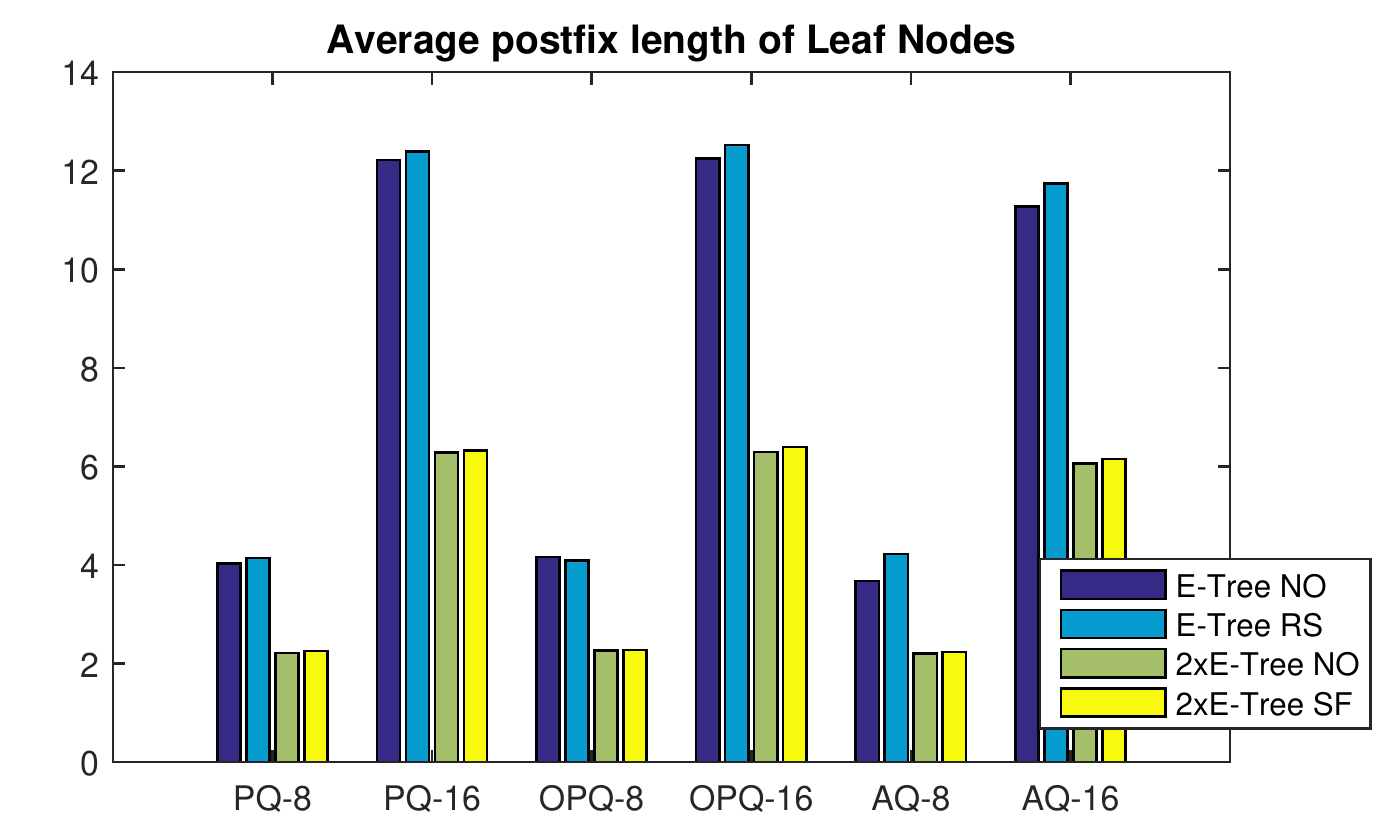}
		\label{depth}
}
\caption{The statistics of Encoding Tree. E-Tree NO(resp. RS) refers to original orderings(resp. randomized orderings) with one E-Tree, 2xE-Tree refers to E-Forest with 2 E-Trees.}.
\label{stat}
\end{figure}

\subsection{Encoding Forests}
We can further lower the calculation time with multiple Encoding Trees. 
An Encoding Forest is generated by splitting the encodings in to several parts and build Encoding Trees separately. 
However an Encoding Tree record every IDs of the dataset vectors on the leaf nodes, 
an Encoding Forest will have to record the vectors IDs for multiple times, resulting in more memory consumption. 

To perform distance calculation with an Encoding Forest consists of multiple Encoding Trees, 
we first calculate the "partial" distance with the two encoding tree and output the result into different arrays. 
Then the distance is obtained by the summation these result arrays. 
Note performing summation of the resulting arrays is time consuming, 
thus it's not recommended to construct an Encoding Forest with too many Encoding Trees. 
As observed in Figure \ref{depth}, we recommend constructing a Encoding Tree with at least 4 codes for speed consideration.

In our implementations, we generate 2 Encoding Trees for a balanced trade-off between memory consumption and calculation time.

\section{Experiments and Discussions}

\begin{figure}[t]
\centering
\subfigure[Distance Computing Time]{
	\includegraphics[width=1\linewidth]{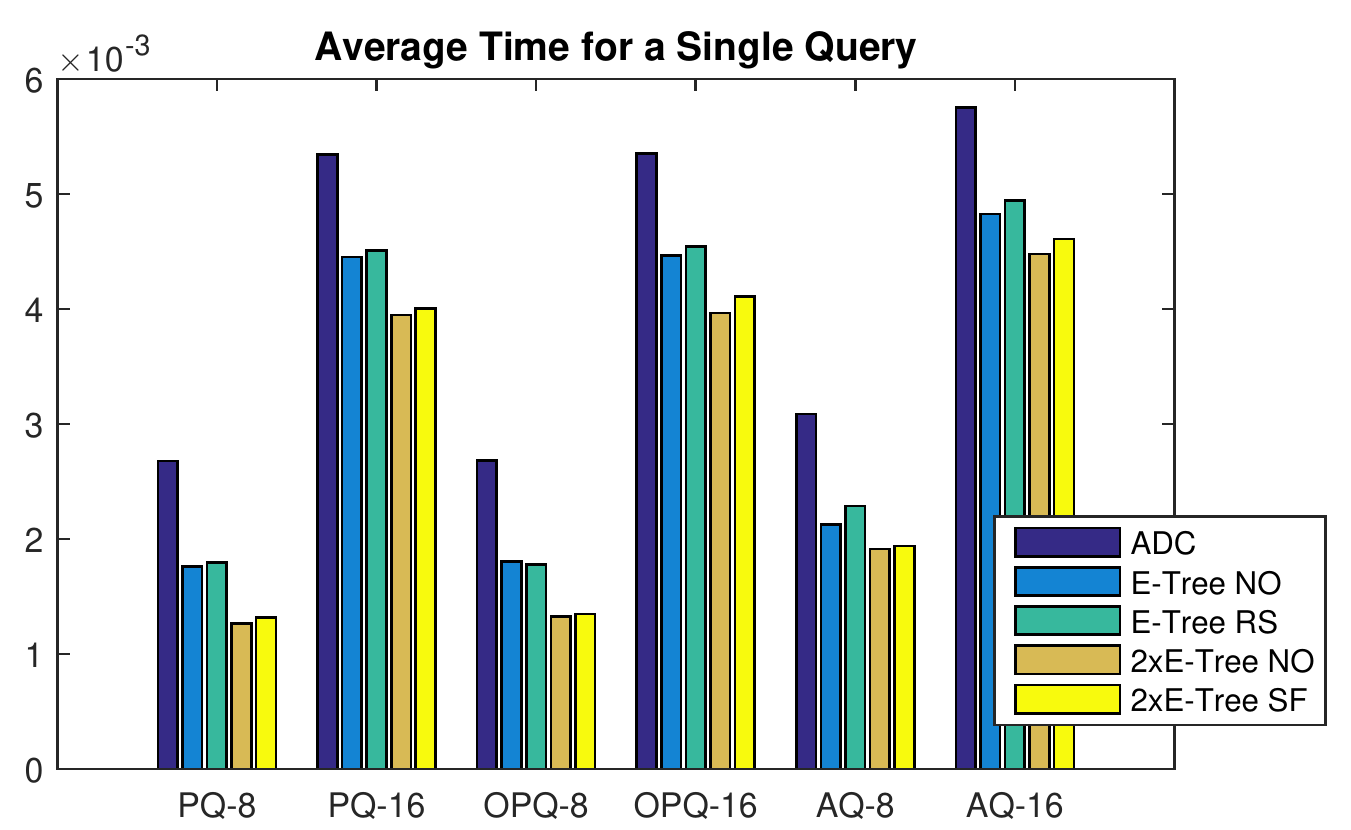}
}
\subfigure[Memory usage for storing the dataset]{
	\includegraphics[width=1\linewidth]{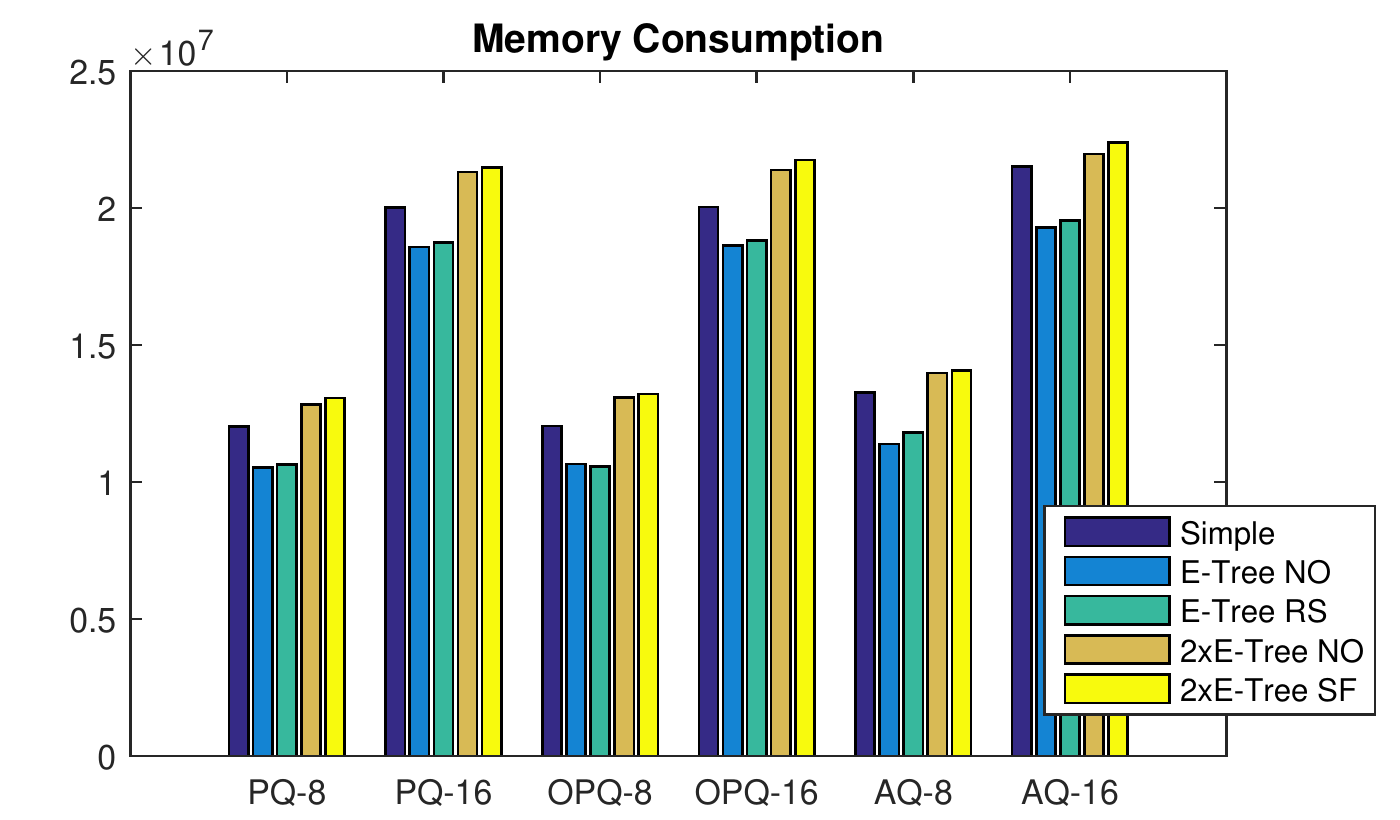}
}
\caption{The performance of accelerated distance computation with Encoding Tree. We include the performance of Asymmetric Distance Computation for reference. }.
\label{perf}
\end{figure}

To examine the acceleration with our method, we generated encoded vectors of SIFT1M dataset with Product Quantization\cite{pq}, 
Optimized Product Quantization\cite{opq} and Additive Quantization\cite{babenko2012inverted}. 
For Additive Quantization we quantized on the extra information to 1 Byte as proposed in \cite{babenko2012inverted} 
and store the extra information on the leaf node. For all methods, we produce $M=8/16, K=256$ encodings. 
The vectors' ID is also stored along with the encoded vectors. 
We used an Core i7 running at 3.6Ghz with 16G memory to perform the experiments.

\subsection{Statistics of Encoding Tree}
There are three things we're interested about the Encoding Tree:
\begin{itemize}
\item Number of internal nodes on the E-Tree.
	Every time we visit and update the distance context table, 
	we're avoiding at least one excessive computation on the dataset (an internal node has at least two children or it is merged to a leaf node)
\item Total number of leaf nodes on the E-Tree.
\item The average postfix length of the leaf nodes.
	The postfixes have to be computed for all leaf nodes and is most time consuming.
\end{itemize}
The statistics is shown in Figure \ref{stat}. 
It can be observed that the number of leaf nodes is almost equal to the the dataset length for a single Encoding Tree, 
this is because vectors are not likely to be encoded into a same encoding. 
Nevertheless, the postfix length is much smaller than encoding length $M$, 
so we can still gain a significant acceleration. 
We also observed the internal nodes are very few compared to the leaf nodes. 
To conclude, most of the time spending on distance computation would be on the postfix computing.

\subsection{Encoding Ordering}
Obviously, the acceleration of distance computation and compression rate of the encoded dataset 
with Encoding Tree is highly dependent on the length of common prefixes. 
If encoded vectors have longer common prefixes, i.e, a deeper depth of the leaf node, 
our proposed encoding tree can perform better. 
The ordering of encoding chunks may have an influence on the final tree size 
and therefore the speed of distance computation. We compared the following ordering of the encodings:
\begin{enumerate}
\item Original ordering. We generate encoding tree directly according to the original encodings.
\item Randomized ordering. We first shuffle the encoding chunks, then generate the encoding tree.
\end{enumerate}
We adopt different encoding arrangement and generate the corresponding E-Tree. 
As depicted in statistics Figure \ref{stat} and performance Figure \ref{perf}. 
We found the encoding orderings have relatively small impact on the number of postfix/prefix length.

\subsection{Performance}
Figure \ref{perf} present the distance computing time and memory consumption for E-Tree and E-Forest. 
E-Tree/E-Forest achieves maximum acceleration ratio on smaller encodings. 
On 8 bytes PQ encoding, the average distance computing time with ADC is \textbf{2.678ms}, 
while it takes only \textbf{1.265ms}(111.7\% speed-up) with E-Forest or \textbf{1.760ms}(52.2\% speed-up) with E-Tree. 
E-Tree also lowers the memory consumption by \textbf{12.5\%}. 
E-Forest achieves very cost effective speed-up with \textbf{6.67\%} more memory consumption.

On longer encodings the postfix length is also increased. 
One may generate E-Forest with more E-Trees on longer encodings to overcome this issue, 
at the cost of increased memory consumption. E-Tree and E-Forest perform best on smaller encodings as the postfix is shorter.

\subsection {Application on related algorithms}
We experiment our methods on two simple utilization of Asymmetric Distance Computation, 
namely, IVFADC proposed in \cite{pq} and Locally Optimized Product Quantization proposed in \cite{kalantidis2014locally}. 
We replace the ADC part with Encoding-Tree to boost the search speed. 
The speed-up is shown in Table \ref{spe}. Similarly, one can apply E-Tree and E-Forrest 
on any circumstance depending on fast approximate distance computation. 
One can also extend E-Tree to allow fast scalar product, etc, we leave it a future work.
\begin{table}
\centering
\renewcommand{\arraystretch}{1.2}
\begin{tabular}{|c||c|c|}
	\hline
	         ~           &        IVFADC        & LOPQ  \\ \hline\hline
	        ADC Time         & (\textit{74ms}) 65ms & 69ms  \\ \hline
	  Time with E-Tree   &         55ms         & 59ms  \\ \hline
	 Time with E-Forest  &         \textbf{42ms}         & \textbf{47ms}  \\ \hline \hline
	       ADC Memory        &        8.01G         & 8.52G \\ \hline
	 Memory with E-Tree  &        \textbf{7.17G}         & \textbf{7.59G} \\ \hline
	Memory with E-Forest &        8.82G         & 9.30G \\ \hline
\end{tabular} 
\caption{Applying E-Tree/E-Forest on IVFADC and LOPQ with configuration $w=64, K'=8192, K=256, M=8$ suggested in \cite{jegou2011searching}. E-Tree brings significant improvement over the original algorithms. Number in brackets are reproduced from \cite{jegou2011searching}. }
\label{spe}
\end{table}

\section{Conclusion}
E-Tree/E-Forest provide significant speed-up and lowers memory consumption by generating a tree to avoid excessive computation. The memory consumption can be also lowered. E-Tree and E-Forest are compatible to current existing algorithms relying on ADC and can bring significant speed-up. In this paper we found the length of postfix is the major limitation of Encoding Tree, how to reduce the length the length of postfix or increase the length of prefix is leave to be the future work.

\bibliographystyle{aaai}
\bibliography{sigproc}  
\end{document}